\documentclass[letterpaper]{article} 
\usepackage{aaai25}  
\usepackage{times}  
\usepackage{helvet}  
\usepackage{courier}  
\usepackage[hyphens]{url}  
\usepackage{graphicx} 
\usepackage{natbib}  
\usepackage{caption} 
\usepackage{algorithm}
\usepackage{algorithmic}
\usepackage{multirow}
\usepackage{colortbl, xcolor} 
\usepackage{booktabs} 
\usepackage{graphicx}
\usepackage{amssymb}
\usepackage{amsmath}
\usepackage{cleveref}

\usepackage{newfloat}
\usepackage{listings}
\DeclareCaptionStyle{ruled}{labelfont=normalfont,labelsep=colon,strut=off} 
\floatstyle{ruled}
\newfloat{listing}{tb}{lst}{}
\floatname{listing}{Listing}
\title{Efficient Event-Based Semantic Segmentation via Exploiting Frame-Event Fusion: A Hybrid Neural Network Approach}
\author{
    Hebei Li\textsuperscript{\rm 1, \dag},
    Yansong Peng\textsuperscript{\rm 1, \dag},
    Jiahui Yuan\textsuperscript{\rm 1},
    Peixi Wu\textsuperscript{\rm 1},
    Jin Wang\textsuperscript{\rm 1} \\
    Yueyi Zhang\textsuperscript{\rm 1, *},
    Xiaoyan Sun\textsuperscript{\rm 1, \rm 2}
}
\affiliations{
    \textsuperscript{\rm 1}MoE Key Laboratory of Brain-inspired Intelligent Perception and Cognition, University of Science and Technology of China\\
    \textsuperscript{\rm 2}Institute of Artificial Intelligence, Hefei Comprehensive National Science Center\\
    \{lihebei, pengyansong, yuanjiahui, wupeixi, jin01wang\}@mail.ustc.edu.cn,
    \{zhyuey, sunxiaoyan\}@ustc.edu.cn
}

\usepackage{bibentry}

\begin{document}

\maketitle
\let\thefootnote\relax\footnotetext{\textsuperscript{\dag}Equal contribution \textsuperscript{*}Corresponding author}
\begin{abstract}
Event cameras have recently been introduced into image semantic segmentation, owing to their high temporal resolution and other advantageous properties. However, existing event-based semantic segmentation methods often fail to fully exploit the complementary information provided by frames and events, resulting in complex training strategies and increased computational costs. To address these challenges, we propose an efficient hybrid framework for image semantic segmentation, comprising a Spiking Neural Network branch for events and an Artificial Neural Network branch for frames. Specifically, we introduce three specialized modules to facilitate the interaction between these two branches: the Adaptive Temporal Weighting (ATW) Injector, the Event-Driven Sparse (EDS) Injector, and the Channel Selection Fusion (CSF) module. The ATW Injector dynamically integrates temporal features from event data into frame features, enhancing segmentation accuracy by leveraging critical dynamic temporal information. The EDS Injector effectively combines sparse event data with rich frame features, ensuring precise temporal and spatial information alignment. The CSF module selectively merges these features to optimize segmentation performance. Experimental results demonstrate that our framework not only achieves state-of-the-art accuracy across the DDD17-Seg, DSEC-Semantic, and M3ED-Semantic datasets but also significantly reduces energy consumption, achieving a 65\% reduction on the DSEC-Semantic dataset.
\end{abstract}
\begin{links}
    \link{Code}{https://github.com/HebeiFast/Hybrid-Segmentation}
\end{links}

\section{Introduction}
    Event cameras \cite{gallego2020event}, also known as neuromorphic cameras, are revolutionizing computer vision by offering a pathway to next-generation performance and efficiency. Unlike traditional frame-based cameras that capture images at fixed intervals, event cameras detect changes in pixel-level intensity asynchronously, focusing exclusively on the dynamic elements of visual scenes \cite{lichtsteiner2008128}. This unique capability provides high dynamic range and low power consumption \cite{gehrig2022high, huang2023event} and makes event cameras particularly well-suited for applications such as autonomous driving, augmented reality, robotic vision, and other real-time processing tasks.

    Among these applications, event-based semantic segmentation (ESS) stands out for its significance in visual perception \cite{alonso2019ev, sun2022ess, li2024event}. However, existing ESS approaches often struggle to fully exploit the complementary information between frames and events. Most methods transfer information from the image domain to the event domain \cite{jing2024hpl, xia2023cmda, kong2024openess}. However, these methods still require complex training strategies and incur significant computational costs due to the reliance on strong encoders. 

    \begin{figure}[!t]
        \centering
        \includegraphics[width=1\linewidth]{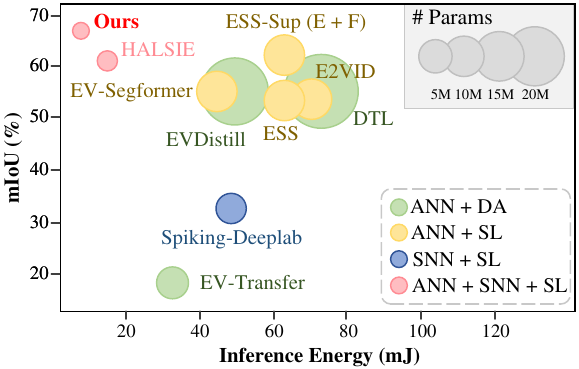}
        \caption{Comparison of various event-based semantic segmentation methods in terms of mIoU (\%) and inference energy (mJ). DA and SL represent ``domain adaptation" and ``supervised learning", respectively.}
        \label{fig:efficient}
        \vspace{-13pt}
    \end{figure}
    
    To address these challenges, Spiking Neural Networks (SNNs) have recently emerged as a promising alternative for processing asynchronous data \cite{sengupta2019going, liao2023convolutional, zhang2019tdsnn, yang2022training}. Unlike traditional artificial neural networks, SNNs operate with spiking neurons that transmit information via discrete spikes. This spiking mechanism offers several inherent advantages, including reduced energy usage, accelerated inference times, and a natural fit for handling event-driven tasks \cite{akopyan2015truenorth, davies2018loihi, pei2019towards, furber2014spinnaker}. Despite these advantages, recent studies that apply SNNs to segmentation tasks \cite{kim2022beyond, patel2021spiking, li2024deep} indicate that there is still substantial room for performance enhancement. While some approaches have attempted to integrate ANN and SNN to leverage their complementary strengths \cite{biswas2024halsie, yue2023hybrid}, these methods often neglect the critical interaction of intermediate layers between temporal information in SNNs and the spatial features extracted by ANNs, leaving significant potential untapped.

    In this work, we propose a novel framework that seamlessly integrates the complementary characteristics of frame and event signals through two neural network branches. The ANN and SNN branches are responsible for feature extraction of the frames and events, respectively. In addition, three specialized modules are developed for the interaction between both branches: the Adaptive Temporal Weighting (ATW) Injector, the Event-Driven Sparse (EDS) Injector, and the Channel Selection Fusion (CSF) module. The ATW Injector is designed to dynamically incorporate temporal features from event data into frame features using adaptive temporal weighting, enhancing the frame features with crucial dynamic temporal information captured by event cameras, which significantly improves segmentation accuracy. The EDS Injector efficiently combines sparse event data with the rich spatial and color features of frames, ensuring precise alignment of temporal and spatial aspects critical for accurate scene understanding in dynamic environments. Additionally, the CSF module selectively integrates key semantic features from both branches, ensuring that only the most essential data contributes to the final segmentation. Experimental results demonstrate that our method achieves state-of-the-art results across the DDD17-Seg, DSEC-Semantic, and M3ED-Semantic datasets, while significantly reducing energy consumption (\Cref{fig:efficient}). Notably, on the DSEC-Semantic dataset, our approach achieves a 65\% reduction in energy consumption compared to the most efficient existing method.
    
    In brief, our contributions are summarized as follows:
    \begin{itemize}
        \item We propose a novel framework that effectively integrates the complementary characteristics of frames and events, leveraging SNN and ANN in two branches to enhance event-based semantic segmentation.
        \item We introduce  three components for branch interaction: the ATW Injector, which integrates dynamic temporal information from event data into the frame stream; the EDS Injector, which aligns sparse event features with frame features; and the CSF module, which selectively merges these features for optimal segmentation.
        \item Experimental results show that our method achieves state-of-the-art performance on multiple datasets, while also significantly reducing energy consumption. 
    \end{itemize}

\section{Related Work}  
\subsection{Event-based Semantic Segmentation}
    The goal of  ESS is to classify pixels of frames into semantic categories with the help of events, thereby enhancing scene understanding. Ev-SegNet \cite{alonso2019ev} pioneered a new representation for event data and introduced the DDD17-Seg dataset specifically for ESS. To reduce latency, HMNet \cite{hamaguchi2023hierarchical} introduced a hierarchical memory-based event encoding approach. To improve accuracy while minimizing the need for extensive event annotations, EvDistil \cite{wang2021evdistill} and DTL \cite{wang2021dual} leveraged knowledge distillation from labeled image data to train networks on unlabeled events. Ev-Transfer \cite{wang2021evdistill} and ESS \cite{sun2022ess} developed unsupervised domain adaptation techniques that adapt existing image datasets to the event domain. Addressing scalability and annotation challenges, OpenESS \cite{kong2024openess} proposed an annotation-free approach by distilling pre-trained vision-language models. More recently, HALSIE \cite{biswas2024halsie} introduced an end-to-end learning framework for ESS that integrates cross-domain learning. In this work, we focus on processing cross-domain data while keeping energy efficiency in ESS.

\subsection{Spiking Neural Network}
    SNNs, often regarded as the third generation of artificial neural networks, simulate biological neurons using binary spiking signals \cite{gerstner2002spiking}. Two main approaches for developing deep SNNs are ANN-to-SNN conversion and direct training. ANN-to-SNN conversion \cite{diehl2015fast, rueckauer2017conversion} translates trained ANNs into SNNs but is limited by the need for large time steps and poor handling of neuromorphic data. Direct training, which better suits temporal data, uses surrogate functions to overcome the non-differentiability of spiking neurons \cite{wu2018spatio}. However, these methods primarily target classification tasks and still underperform compared to ANNs, especially in complex tasks such as semantic segmentation.
    
    To address the limitations of current models, several studies \cite{liu2023motion, biswas2024halsie} have investigated the combination of ANNs and SNNs. A notable approach employs an SNN for the encoder and an ANN for the decoder \cite{kugele2021hybrid, zhang2021event}. This configuration optimizes performance by leveraging the precise temporal dynamics of SNNs while utilizing the robust representational capabilities of ANNs for reconstruction tasks. Additionally, multi-modal strategies \cite{lee2022fusion, zhang2024accurate} have been explored where RGB data is processed by ANNs and event data by SNNs, facilitating cross-modal fusion. This method significantly improves the system’s ability to interpret complex scenes with greater accuracy. Despite these advancements, most existing methods do not fully harness the potential of SNNs during feature extraction, often treating them merely as auxiliaries to ANNs without true integration. In this work, we aim to effectively integrate the continuous activations of ANNs with the discrete spikes of SNNs, thereby enhancing joint processing capabilities in complex perceptual tasks.
 
\begin{figure*}[t]
    \centering
    \includegraphics[width=1\textwidth]{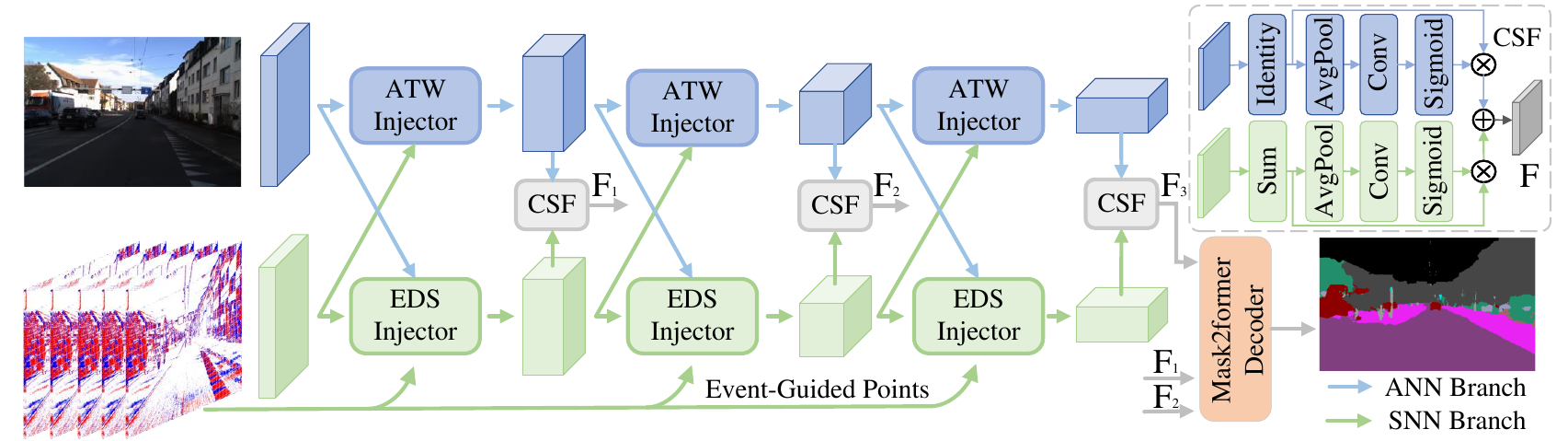}
    \caption{Overview of the proposed framework.The architecture integrates ANN and SNN branches for efficient semantic segmentation. The ATW Injector injects important event-based features within the ANN branch, while the EDS Injector enhances SNN features with color information from the ANN branch. The CSF module selectively fuses features from both branches. These multi-scale features from the CSF module are then fed into the Mask2Former Decoder.}
    \label{fig:Arch}
    \vspace{-10pt}
\end{figure*}

\section{Preliminary}
The original event stream comprises a sequence of 4-tuples $E = \{e_k\}_{k=1}^{N_e}$, where $N_e$ denotes the total count of events. Each event $e_k$ contains four attributes: $(x_k, y_k, t_k, p_k)$, which represent the spatial coordinates, timestamp, and  the polarity of brightness change, respectively. 

In this work, we transform asynchronous event streams into corresponding voxel grids \cite{gehrig2019end}, represented as $V\in \mathbb{R}^{B\times H\times W}$, where $B$, $H$, and $W$ denote the number of time bins, height, and width of the grids, respectively. Each time bin within the voxel grid is defined as
\begin{equation}
    V(b) = \sum_i p_i \max(0, 1-|b-\frac{t_i - t_{start}}{t_{end} - t_{start}}(B-1)|),
\end{equation}
where $b \in \{0,\cdots, B-1\}$ represents the $b$-th time bin. $t_{start}$ and $t_{end}$ indicate the start and end timestamps of the event stream, respectively.
To ensure uniformity in scale and mitigate the impact of outliers, we normalize the voxel grid $V$ using the Z-score normalization. 

\begin{figure}[!t]
    \centering
    \includegraphics[width=1\linewidth]{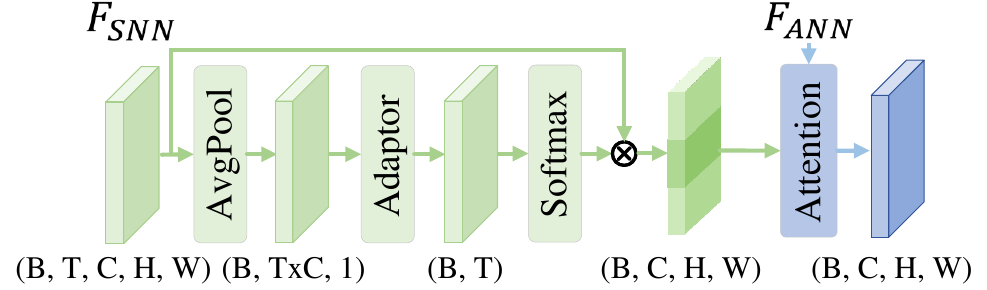}
    \caption{Illustration of the proposed Adaptive Temporal Weighting (ATW) Injector.}
    \label{fig:AWT}
    \vspace{-10pt}
\end{figure}

\section{Methodology}
\subsection{Main Architecture}
The architecture of our method, illustrated in \Cref{fig:Arch}, is built upon a novel frame-event fusion strategy within a hybrid neural network framework. This design integrates both ANN and SNN, leveraging the distinct advantages of conventional frame-based and event-based data. The architecture consists of two parallel branches: the ANN branch, which processes frame data using Adaptive Temporal Weighting (ATW) Injectors, and the SNN branch, which processes event data via Event-Driven Sparse (EDS) Injectors. Both branches converge in Channel Selection Fusion (CSF) modules, which extract the most relevant semantic features from each branch. These multi-scale features from the CSF modules are then fed into the Mask2Former Decoder, yielding the final segmentation output.

\subsection{Adaptive Temporal Weighting Injector}
The Adaptive Temporal Weighting (ATW) Injector, as illustrated in \Cref{fig:AWT}, is designed to dynamically integrate SNN features into ANN features through cross attention. The primary objective is to leverage temporal information to optimize spatial feature representations. This integration not only improves segmentation accuracy in rapidly changing environments but also injects important event-based features into the ANN representation.
\begin{figure}[!t]
    \centering
    \includegraphics[width=1\linewidth]{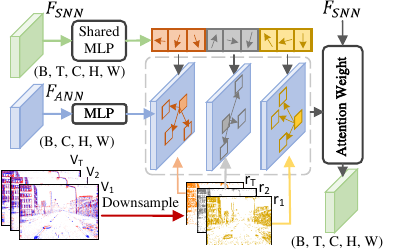}
    \caption{Illustration of the proposed Event-driven Sparse (EDS) Injector.}
    \label{fig:EDS}
    \vspace{-10pt}
\end{figure}
The ATW initially receives an SNN feature from event data $F_{SNN} \in \{0,1\}^{B\times T\times C\times H\times W}$. Considering the sparse nature of SNN features, we apply an average pooling operation to the spatial dimensions of ${F}_{SNN}$ to effectively capture the overall sparsity across the $T\times C$, resulting in the squeezed feature ${F}_{SNN}^{Pool}  \in \mathbb{R}^{B\times T \times C}$.
Subsequently, we further utilize the Adaptor~\cite{houlsby2019parameter} to capture the global information along temporal and channel dimensions while maintaining low computational overhead. The adapter contains a downsample and an upsample. The output of the adaptor is then averaged across the channel dimension $C$ to produce a temporal weighting vector $W_{SNN} \in \mathbb{R}^{B\times T \times C}$
\begin{equation}
    W_{SNN} = W_{up}^T \sigma(W_{down}^T {F}^{Pool}_{SNN}),
\end{equation}
where $W_{down}$ and $W_{up}$ are the weights applied in the downsample and upsample, respectively. $\sigma$ is the ReLU activation function. 
Next, the dynamic weighted SNN features are obtained through the operation of normalizing the resulting temporal weights and multiplying them with the original SNN input tensor $F_{SNN}$, shown as
\begin{equation}
   F_{SNN}^W = \sum_t^T \alpha_{t,c} \cdot F_{SNN},
\end{equation}
where $\alpha_{t,c}$ is the output of softmax function on $W_{SNN}$ along channel and temporal dimension. Finally, this weighted tensor $F_{SNN}^W$ is integrated with the ANN feature $F_{ANN}$ via sparse attention layer \cite{zhudeformable}, described as
\begin{equation}
   F_{ANN}^o = Attention(F_{SNN}^W, F_{ANN}),
\end{equation}
where $F_{ANN}^o \in \mathbb{R}^{B\times C \times H \times W}$ is the optimized output feature map used in the subsequent stages of the ANN branch.

\subsection{Event-driven Sparse Injector}
The Event-driven Sparse (EDS) Injector, as illustrated in \Cref{fig:EDS}, is designed to enhance the temporal feature extraction capabilities of SNN features by integrating them with the rich color information from ANN features derived from image data. 
By injecting detailed visual information into the sparse SNN features, the module enhances the SNN branch’s ability to efficiently and accurately extract temporal features, enabling a deeper understanding of complex dynamic scenes.

The EDS Injector module receives three distinct inputs: an SNN feature ${F}_{SNN}$, an ANN feature ${F}_{ANN}$ and a voxel grid $V$. The SNN feature ${F}_{SNN}$, containing temporal changes in the scene, initially passes through a Shared MLP layer, which generates directional offsets for each time step 
$T$. These offsets adjust the spatial positioning of the temporal features. The resulting directional offsets are denoted as
\begin{equation}
    \Delta r = ShareMLP({F}_{SNN}),
\end{equation}
where $\Delta r \in \mathbb{R}^{B\times T \times N_{offset} \times N_{points}}$ represents the different offset of the direction for sample points. 

Previous methods \cite{zhudeformable} use dense reference points during feature extraction, resulting in significant computational overhead for large feature maps. However, the sparsity of event signals does not require dense sampling. We leverage the sparse nature of event signals as guided reference points $r$ to reduce computational costs. To obtain the sparse reference points $r$, we downsample the event voxel $V$ to align the spatial resolution of $F_{ANN}$ and acquire the non-zero values from the voxel grid as reference points. At the same time, the ANN features pass through an MLP layer, aligning them with the SNN feature space. The projected ANN features ${F}_{ANN}^{Proj}$ are denoted as
\begin{equation}
    {F}_{ANN}^{Proj} = MLP({F}_{ANN}).
\end{equation}
The projected ANN features ${F}_{ANN}^{Proj}$ are then combined with the directional offsets $\Delta r$ generated by the SNN branch. This combination uses deformable attention \cite{zhudeformable} with event-guided reference points. The deformable attention mechanism adjusts the sampling locations based on the directional offsets, allowing the network to integrate rich color information into SNN features
\begin{equation}
    F_{SNN}^o = \sum_{k=1}^K A_{SNN} \cdot (F_{ANN}^{Proj} \cdot F_{SNN}[r_k + \Delta r_k]),
\end{equation}
where $A_{SNN}$ and $F_{ANN}^{Proj}$ are the weighting coefficient and the projected ANN feature map, respectively. $r_k$, $\Delta r_k$ and $K$ are the reference positions from events, learned offset, and the number of sampling points, respectively. $F_{SNN}^o \in \mathbb{R}^{B \times T \times C \times H \times W}$ is the output feature of the EDS injector, delivered to subsequent stages of the SNN branch.

\begin{table*}[t]
    \centering
    \renewcommand{\arraystretch}{1.3} 
    \resizebox{\linewidth}{!}{ 
    \begin{tabular}{lccccccc}
    \hline\hline
    \textbf{Methods} & \textbf{Acc.[\%]} & \textbf{mIoU[\%]} & \textbf{Network} & \textbf{Para.(M)} & \textbf{\#GFLOPs\(_{ANN}\)} & \textbf{\#GFLOPs\(_{SNN}\)} & \textbf{E\(_{Total}\)(mJ)} \\
    \midrule
    EV-SegNet \cite{alonso2019ev}      & 89.76 & 54.81 & ANN    & 29.09 & 73.62 & -     & 338.65   \\
    EVDistill \cite{wang2021evdistill}       & -     & 58.02 & ANN    & 59.34 & 12.45 & -     & 57.27    \\
    DTL \cite{wang2021dual}           & -     & 58.80 & ANN    & 60.48 & 16.74 & -     & 77.01    \\
    Spiking-Deeplab \cite{kim2022beyond} & -     & 33.70 & SNN    & 4.14  & -     & 54.35 & 48.91    \\
    E2VID \cite{rebecq2019high}           & 87.91 & 57.32 & ANN    & 10.71 & 16.65 & -     & 76.59    \\
    EV-Transfer \cite{messikommer2022bridging}     & 47.37 & 14.91 & ANN    & 7.37  & 7.88  & -     & 36.25    \\
    ViD2E        \cite{gehrig2020video}   & 90.19 & 56.01 & ANN    & 29.09 & 73.62 & -     & 338.65   \\
    ESS           \cite{sun2022ess}  & 88.43 & 53.09 & ANN    & 12.91 & 14.22 & -     & 65.41    \\
    ESS-Sup (E)    \cite{sun2022ess} & 91.08 & 61.37 & ANN    & 12.91 & 14.22 & -     & 65.41    \\ 
    ESS-Sup (E+F)  \cite{sun2022ess} & 90.37 & 60.43 & ANN    & 12.91 & 14.22 & -     & 65.41    \\
    EV-Segformer \cite{jia2023event}    & 94.72 & 54.41 & ANN    & 44.61  & 9.88     & -     & 45.42        \\
    OpenESS       \cite{kong2024openess}  & 91.05 & 63.00 & ANN    & -     & -     & -     & -        \\
    HALSIE        \cite{biswas2024halsie}  & 92.50 & 60.66 & Hybrid & 1.82  & 3.84  & 0.267 & 17.89    \\
    \hline
    Ours            & \textbf{95.07} & \textbf{67.31} & Hybrid & \textbf{1.79}  & \textbf{1.95}     & \textbf{0.110}     & \textbf{9.08}  \\
    \hline\hline
    \end{tabular}
    }
    \caption{Quantitative segmentation results on the DDD17-Seg dataset.}
    \label{tab:DDD17}
    \vspace{-5pt}
\end{table*}

\begin{table}[!t]
    \centering
    \renewcommand{\arraystretch}{1.0} 
    \scriptsize
    \begin{tabular}{lcccc}
    \hline\hline
    \multicolumn{5}{c}{DSEC-Semantic}\\
    \midrule
    \textbf{Methods} & \textbf{Acc.[\%]} & \textbf{mIoU[\%]} & \textbf{Para.(M)} & \textbf{E\(_{Total}\)(mJ)} \\
    \midrule
    EV-SegNet       & 88.61 & 51.76 & 29.09 & 1863.83 \\
    E2VID         & 87.55 & 49.66 & 10.71 & 416.99  \\
    EV-Transfer    & 60.50 & 23.20 & 7.37  & 197.48  \\
    ESS         & 84.16 & 45.38 & 12.91 & 356.32  \\
    ESS-Sup (E)     & 89.25 & 51.57 & 12.91 & 356.32  \\
    ESS-Sup (E+F) & 89.37 & 53.29 & 12.91 & 356.32  \\
    OpenESS     & 90.21 & 57.21 & -     & -       \\
    HMNet-B1     & 88.94 & 51.42 & 4.07  & 1974.50 \\
    HMNet-L1      & 89.89 & 55.09 & 5.51  & 6686.38 \\
    HALSIE       & 89.01 & 52.43 & 1.82  & 94.41   \\
    \hline
    \textbf{Ours}   & \textbf{94.27} & \textbf{66.57} & \textbf{1.79} & \textbf{33.45} \\
    \hline\hline
    \multicolumn{5}{c}{M3ED-Semantic}\\
    \midrule
    E2VID           & 78.65 & 39.40 & 10.71 & 764.33 \\
    ESS-Sup (E)    & 75.97 & 34.82 & 12.91 & 471.56 \\
    ESS-Sup (E+F)  & 76.40 & 35.19 & 12.91 & 471.56 \\
    \hline
    \textbf{Ours}   & \textbf{87.31} & \textbf{50.91} & \textbf{1.79} & \textbf{70.98} \\
    \hline\hline
    \end{tabular}
    \caption{Quantitative segmentation results on the DSEC-Semantic dataset and M3ED-Semantic dataset.}
    \label{tab:other}
\end{table}

\subsection{Channel Selection Fusion}
The Channel Selection Fusion (CSF) module, as depicted in \Cref{fig:Arch}, aims to effectively combine features from both the ANN and SNN branches to enhance the representation of critical semantic information. Inspired by \cite{hu2018squeeze}, the CSF module emphasizes selecting the most informative channels from these two distinct feature sources. 

The CSF module initially processes the $F_{ANN}^o$ through an identity function to obtain $x_{ANN}$ and the ${F}_{SNN}^o$ through a sum operation to obtain $x_{SNN}$, respectively. After that, each branch utilizes the same operation flow for $x_{ANN}$ and $x_{SNN}$, denoted as 
\begin{equation}
    s = x \cdot AvgPool(Conv(Sigmoid(x)))
\end{equation}
where $s$ and $x$ represent the selected channel features and the input feature map from either the ANN or SNN branch.
Following this, the weighted feature maps $s_{ANN}$ and $s_{SNN}$ are summed to produce the fused feature map $F \in \mathbb{R}^{B\times C \times H \times W}$. Through the channel selection and weighted fusion strategy, the CSF module highlights the feature channels by suppressing irrelevant or redundant features. Finally, the multi-scale outputs of the CSF module are sent to the mask2former decoder.

\section{Experiments and Evaluation}
\subsection{Experimental Setup}
For the implementation of our network, we incorporate frameworks from Detectron and SpikingJelly to establish a robust system for ESS. To facilitate smoother data handling and integration, all datasets in our study are standardized to a consistent format, ensuring consistency in data reading and loading. We train our model for 50000 iterations using the AdamW optimizer alongside the WarmupPolyLR scheduler. We also utilize AMP and gradient clipping techniques to optimize the network efficiently. For the SNN branch, we use the LIF model and the sigmoid function as surrogate gradients to approximate the gradients of the Heaviside function during backpropagation. The timestep of the SNN is 5. The tau and threshold parameters of LIF spiking neurons are 2.0 and 1.0, respectively. For event representations, the event window between consecutive frames is discretized into B = 5 temporal bins. These event representations and synchronized frames are processed by the ANN and SNN branches, respectively. In our experiments, we report accuracy and mean intersection over union (mIoU) on our segmentation maps to evaluate performance. We also estimate energy costs by calculating the average number of FLOPs for each dataset. Further details on the energy estimation process are found in the supplementary material. All experiments use 8 NVIDIA A800 GPUs.

\subsection{Evaluation on DDD-17-Seg Dataset}
\subsubsection{Dataset and Training Details:}
DDD17-Seg~\cite{alonso2019ev} is a well-established ESS benchmark dataset that consists of 40 sequences captured using a DAVIS346 camera along with synchronized grayscale frames, events and segmentation masks. The dataset includes a total of 15,950 training and 3,890 testing event sequences, each with a spatial resolution of 346 × 200. For each time step, we convert the 6400 events into an event voxel grid. To enhance the diversity of the training data, we apply data augmentation techniques such as resizing, cropping, and random flipping.

\subsubsection{Results:}
\Cref{tab:DDD17} presents a comprehensive performance comparison of our method against several state-of-the-art approaches on the DDD17-Seg dataset. Our approach outperforms all competing methods, achieving an accuracy of 95.07\% and a mIoU of 67.31\%, which are the highest among the methods evaluated. Notably, our model utilizes an efficient network architecture, which significantly reduces the number of parameters to 1.79 million, and requires only 1.95 GFLOPs for the ANN component and 0.110 GFLOPs for the SNN component. This efficiency leads to a total energy consumption of just 9.08 mJ, making our method not only accurate but also computationally efficient. This demonstrates the effectiveness of our model in leveraging the strengths of both ANN and SNN architectures.

\begin{figure*}[!t]
    \centering
    \includegraphics[width=1\linewidth]{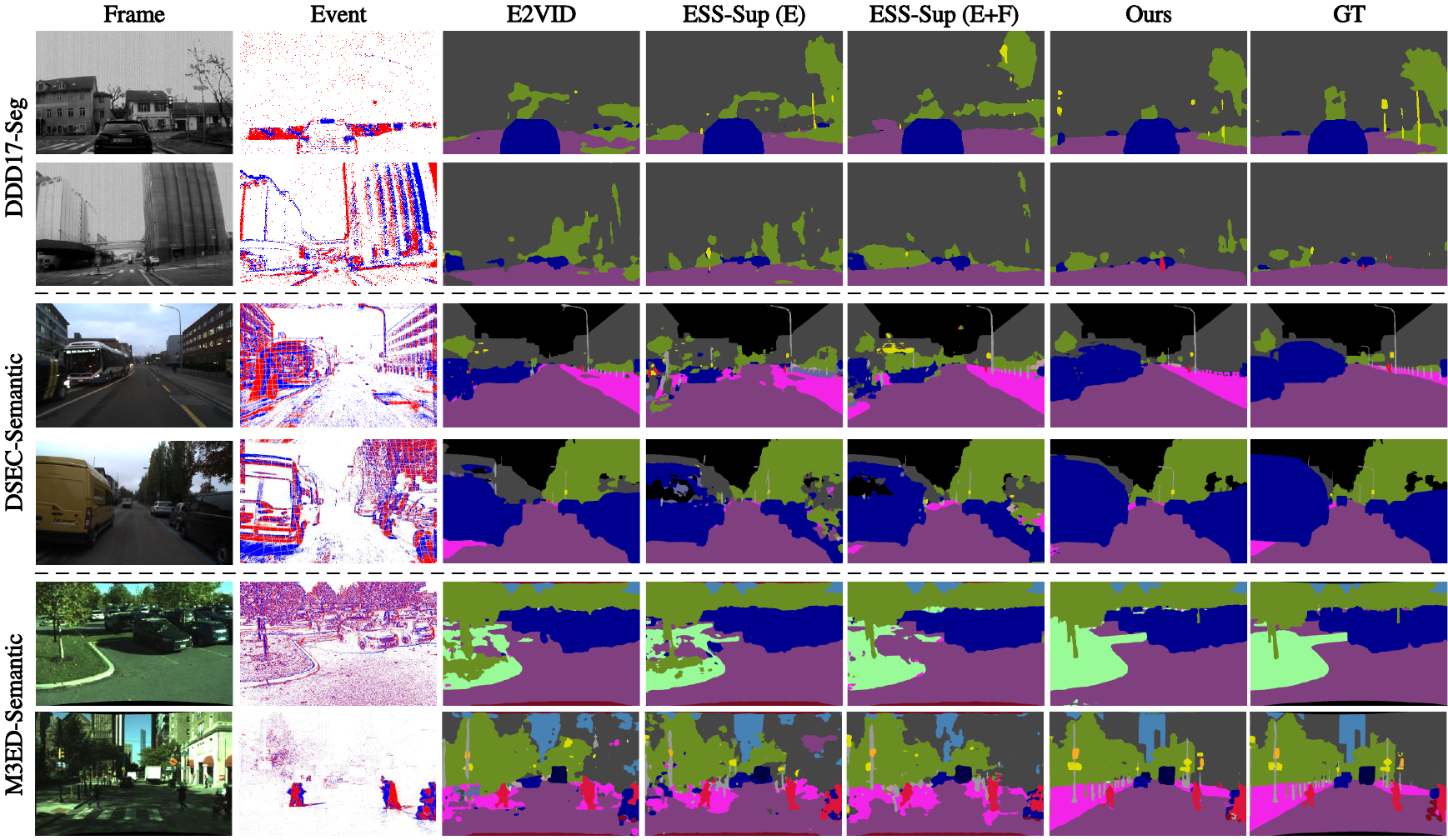}
    \caption{Qualitative segmentation results from the DDD17-Seg, DSEC-Semantic, and M3ED-Semantic datasets.}
    \label{fig:vis}
\end{figure*}

Rows 1-2 of \Cref{fig:vis} provide qualitative comparisons of segmentation results on the DDD17-Seg dataset. Our approach consistently outperforms other methods, particularly in accurately segmenting challenging regions with complex driving scenes. The comparison between our method against E2VID and ESS-Sup (E) demonstrates that our method effectively captures finer details and exhibits improved segmentation accuracy, particularly in areas with rapid motion and fine structures. The visual results validate the efficacy of our framework in leveraging event-based information to enhance segmentation performance. Additionally, our method is able to capture semantic information about people that is not annotated in the ground truth.

\subsection{Evaluation on DSEC-Semantic Dataset}
\subsubsection{Dataset and Training Details:}
The DSEC-Semantic dataset~\cite{sun2022ess} extends the DSEC dataset by providing semantic labels for 11 sequences. This dataset includes 4017 training and 1395 testing event sequences, each with a spatial resolution of 640 × 440. Different from DDD17-Seg, the number of selected events for each event voxel is 100000.


\subsubsection{Results:}
\Cref{tab:other} provides a detailed comparison of our method against various state-of-the-art approaches on the DSEC-Semantic dataset. Our method achieves the highest accuracy at 94.27\% and a mIoU of 66.57\%, outperforming all other methods. Despite its superior performance, our model is highly efficient, requiring only 1.79 million parameters, which is significantly lower than most other approaches. Additionally, our approach has a remarkably low energy consumption of 33.45 mJ, demonstrating the efficiency of our method in terms of energy resources. This efficiency is particularly notable when compared to methods such as EV-SegNet and HMNet-L1, which consume much higher energy levels. Overall, our approach balances high performance with low energy consumption, making it a robust solution for ESS tasks on the DSEC-Semantic dataset.

Rows 3-4 of \Cref{fig:vis} provide qualitative comparisons of segmentation results on the DSEC-Semantic dataset. Our method demonstrates superior performance in capturing fine details and accurately segmenting complex urban scenes, particularly in areas with dynamic objects and challenging lighting conditions. Compared to other methods, our approach better aligns with the ground truth, effectively combining the temporal and spatial information to enhance semantic segmentation accuracy.

\subsection{Evaluation on M3ED-Semantic Dataset}
\subsubsection{Dataset and Training Details:}
The M3ED-Semantic dataset, which is proposed in our paper, is a subset of the M3ED dataset~\cite{chaney2023m3ed} that provides large-scale high-quality, synchronized data from a variety of robotic platforms operating in diverse and challenging environments. We select 19 sequences, with 5,306 training sequences and 1,969 testing sequences for constructing the M3ED-Semantic dataset. Each sequence includes corresponding images, events, and segmentation masks annotated with 18 distinct labels. All data are captured at the resolution of 960$\times$640. The other setting is the same as the DSEC-Semantic dataset. The details description is provided in the supplementary material. 

\Cref{tab:other} presents a performance comparison of various methods on the M3ED-Semantic dataset, demonstrating the superiority of our approach in ESS. Our method achieves an accuracy of 87.31\% and a mIoU of 50.91\%, which are significantly higher than those of other competing methods. Specifically, E2VID records an accuracy of 78.65\% and a mIoU of 39.40\%, while the ESS-Sup (E) and ESS-Sup (E+F) methods report lower accuracies and mIoUs under similar conditions. Our method demonstrates remarkable energy efficiency with a total energy consumption of only 70.98 mJ, considerably less than the 764.33 mJ required by E2VID and 471.56 mJ by both ESS-Sup configurations. Overall, our method achieves high performance while maintaining low computational and energy costs.

Rows 5-6 of \Cref{fig:vis} show the segmentation results from various methods on the M3ED-Semantic dataset. The results are illustrated for scenes captured from both drone and vehicle-mounted cameras. Our method exhibits superior segmentation performance, particularly in distinguishing between different classes in complex outdoor environments, such as roads, buildings, and vegetation. Compared to other methods, our approach demonstrates better alignment with the ground truth, accurately segmenting key elements in the scene even under challenging conditions like varying viewpoints and lighting. The integration of event data with frame-based features significantly enhances the model's ability to capture fine details, leading to more precise segmentation in dynamic and diverse scenarios.

\begin{table}[!t]
\footnotesize
    \centering
    \renewcommand{\arraystretch}{1.0} 
    \begin{tabular}{ccccccc}
    
    \hline\hline
    \textbf{\#} & \textbf{F-E} & \textbf{AWT} & \textbf{EDS} & \textbf{CSF} & \textbf{Acc.[\%]} & \textbf{mIoU[\%]} \\
    \midrule
    1    & \checkmark &  &  &  & 92.82 & 61.06 \\
    2    & \checkmark & \checkmark & & & 93.23 & 63.59 \\
    3    & \checkmark & & \checkmark & & 93.05 & 63.71 \\
    4    & \checkmark & & & \checkmark & 93.01 & 62.98 \\
    5    & \checkmark & \checkmark & \checkmark & & 93.39 & 64.37 \\
    6    & \checkmark & \checkmark & & \checkmark & 93.58 & 64.18 \\
    7    & \checkmark & & \checkmark & \checkmark & 93.60 & 64.36 \\
    8    & \checkmark & \checkmark & \checkmark & \checkmark & \textbf{94.27} & \textbf{66.57} \\
    \hline\hline
    \end{tabular}
    \caption{Performance of models with  different configurations on the DSEC-Semantic dataset.}
    \label{tab:ablation_1}
\end{table}

\begin{table}[!t]
    \centering

    \renewcommand{\arraystretch}{1.0} 
    \resizebox{\linewidth}{!}{
        \begin{tabular}{lccc}
        \hline\hline
        \textbf{Fusion} & \textbf{Acc.[\%]} & \textbf{mIoU[\%]} & \textbf{Para. (M)} \\
        \midrule
        EOLO  \cite{cao2024chasing} & 93.39 & 64.39 & 4.50 \\
        REF  \cite{zhou2023rgb}  & 93.53 & 65.14 & 2.98 \\
        CF   \cite{jiang2024complementing}  & 93.49 & 64.19 & 6.32 \\ 
        Ours   & \textbf{94.27} & \textbf{66.57} & \textbf{1.79} \\
        \hline\hline
        \end{tabular}
    }
    \caption{Quantitative results of different fusion methods on the DSEC-Semantic dataset.}
    \label{tab:ablation_fusion}
   
\end{table}

\subsection{Ablation Study}
\subsubsection{The Effectiveness of AWT, EDS, and CSF Modules.}
\Cref{tab:ablation_1} presents the impact of different module configurations on the performance of our model. The baseline model, consisting of frame and event branches, achieves an accuracy of 92.82\% and a mIoU of 61.06\%. When the AWT module is added to the baseline, the performance significantly improves, with accuracy rising to 93.23\% and mIoU to 63.59\%. This improvement underscores the importance of dynamically integrating temporal information into the RGB feature extraction process. Further enhancement is observed with the inclusion of the EDS module alongside AWT, resulting in an accuracy of 93.39\% and mIoU of 64.37\%. The EDS module effectively refines the event data processing by focusing on dynamically sampled features from RGB data, which contributes to the overall accuracy and segmentation quality. The addition of the CSF module, which refines feature representations on a channel-wise basis, achieves the highest accuracy of 94.27\% and mIoU of 66.57\%. 

\subsubsection{Evaluation of Fusion Strategies.}
\Cref{tab:ablation_fusion} evaluates the effectiveness of different fusion strategies in our model. The EOLO method achieves an accuracy of 93.39\% and a mIoU of 64.39\%, with a parameter count of 4.50M. REF achieves slightly higher accuracy and mIoU scores at 93.53\% and 65.14\% with a reduced parameter count of 2.98M, indicating a more efficient use of parameters. The CF method, while comparable in accuracy to EOLO, slightly underperforms in mIoU and requires a significantly higher number of parameters. Our proposed fusion method outperforms all others  while using only 1.79M parameters, the lowest among all the methods compared. This result demonstrates the superior efficiency and effectiveness of our fusion strategy, which improves performance and  significantly reduces the complexity in terms of parameters.

\subsubsection{Evaluation of Timesteps.} \Cref{tab:timestep} illustrates the impact of varying the number of timesteps on our model's performance. Initially, using a single timestep results in an accuracy of 93.69\% and a mIoU of 64.90\%. As the number of timesteps increases to 3, there are slight improvements, with accuracy rising to 93.91\% and mIoU to 65.45\%. The optimal performance is achieved with 5 timesteps. This indicates that the model benefits from a balanced integration of temporal information in this configuration. However, when the timesteps are further increased to 7, both accuracy and mIoU show a slight decline, suggesting that the model may struggle to effectively utilize excessive temporal data, which could introduce noise or increase computational complexity, ultimately limiting performance.

\begin{table}[!t]
    \centering
    \renewcommand{\arraystretch}{1.0} 
    \footnotesize
    \begin{tabular}{lcccc}
    \hline\hline
    \textbf{Timesteps} & \textbf{1} & \textbf{3} & \textbf{5} & \textbf{7} \\
    \midrule
    \textbf{Accuracy [\%]} & 93.69 & 93.91 & \textbf{94.27} & 93.88\\
    \textbf{mIoU [\%]}     & 64.90 & 65.45 & \textbf{66.57} &  65.14 \\
    \hline\hline
    \end{tabular}
    \caption{Performance of models at different timesteps on the DSEC-Semantic dataset.}
    \label{tab:timestep}
\end{table}

\section{Conclusion}
In this paper, we introduced a novel hybrid neural network framework that integrates ANN and SNN to enhance the performance and efficiency of event-based semantic segmentation tasks. By leveraging the complementary strengths of both ANNs and SNNs, our framework effectively processes frame-based and event-based data, addressing the challenges of existing event-based semantic segmentation methods. The key components of our framework include the ATW Injector, which optimizes temporal feature extraction; the EDS Injector, which aligns and integrates sparse event data with rich spatial and color features; and the CSF module, which selectively merges features from both branches. Experimental results across multiple datasets demonstrate that our method achieves state-of-the-art segmentation accuracy and significantly reduces energy consumption. This work paves the way for more advanced hybrid models that can further optimize the synergy between frame-based and event-based processing for various computer vision tasks.

\section{Acknowledgments}
This work was in part supported by the National Natural Science Foundation of China under grants 62472399 and 62021001.

\bigskip

\bibliography{aaai25}

\end{document}